\documentclass[10pt,twocolumn,letterpaper]{article}

\usepackage{cvpr}              

\usepackage{multirow}
\usepackage{tabu}
\usepackage{lineno}

\definecolor{cvprblue}{rgb}{0.21,0.49,0.74}

\usepackage[pagebackref,breaklinks,colorlinks,allcolors=cvprblue]{hyperref}
\usepackage[ruled,vlined]{algorithm2e}

\usepackage{amsmath}
\usepackage{mathrsfs}
\usepackage{bbding}
\usepackage{graphicx}
\usepackage{pifont}
\usepackage{natbib}
\usepackage{colortbl}
\usepackage{cuted}
\usepackage{xcolor}

\SetKwComment{Comment}{// }{}

\def\figurePath{fig/}

\def\mycfiguret#1#2#3{\begin{figure*}[htb]\centering\includegraphics*[clip, width = \linewidth]{\figurePath#2}\vspace{-0.2cm}\caption{#3}\label{fig:#1}\end{figure*}}

\newcommand{\hanxiao}[1]{\textcolor{black}{#1}}

\newcommand{\secref}[1]{Sec.~\ref{#1}}
\renewcommand{\eqref}[1]{Eqn.~(\ref{#1})}

\def\paperID{2668} 
\def\confName{CVPR}
\def\confYear{2025}

\title{SVG-IR: \textbf{S}patially-Varying \textbf{G}aussian Splatting for \textbf{I}nverse \textbf{R}endering}

\newcommand\blfootnote[1]{%
  \begingroup
  \renewcommand\thefootnote{}\footnote{#1}%
  \addtocounter{footnote}{-1}%
  \endgroup
}

\author{Hanxiao Sun$^{1}$ \quad Yupeng Gao$^{2}$ \quad Jin Xie$^{2}$ \quad Jian Yang$^{1}$ \quad Beibei Wang$^{2}$$^{*}$   \vspace{0.3em} \\
{\normalsize $^1$Nankai University} \quad
{\normalsize $^2$Nanjing University} \quad 
}

\begin{document}
\maketitle

\begin{strip}
    \centering
    \vspace{-40px}
    \includegraphics[width=1.0\textwidth]{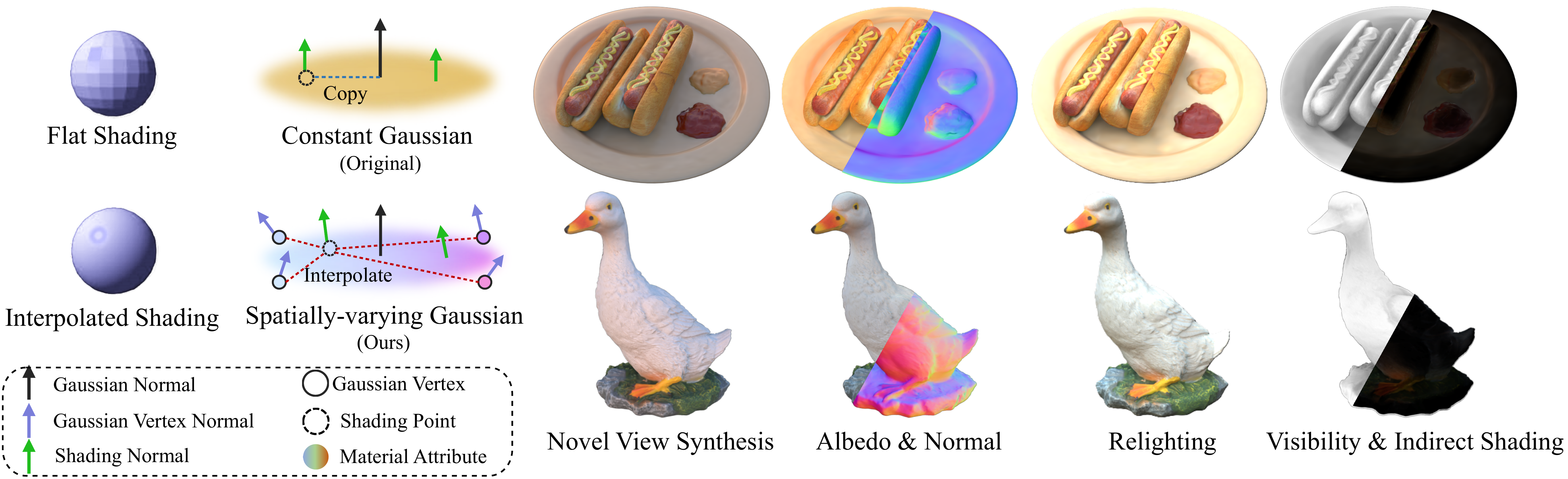}
    \vspace{-20px}
    \captionof{figure}{
 SVG-IR introduces a new spatially-varying Gaussian representation inspired by replacing flat shading for interpolated shading in triangle rendering. Each spatially-varying Gaussian allows spatially-varying material and normal attributes by interpolating among Gaussian vertices defined in the tangent space of a Gaussian. Compared to the original Gaussian (i.e., Constant Gaussian) with constant attributes, SVG has a more powerful representation ability to produce high-quality NVS and relighting results.
    }
    \label{fig:fig_teaser}
    \vspace{-3mm}
\end{strip}

\blfootnote{* Corresponding author.}
\blfootnote{$^1$ College of Computer Science, Nankai University, Tianjin, China}
\blfootnote{$^2$ School of Intelligence Science and Technology, Nanjing University, Suzhou, China}

\begin{abstract}
Reconstructing 3D assets from images, known as inverse rendering (IR), remains a challenging task due to its ill-posed nature. 3D Gaussian Splatting (3DGS) has demonstrated impressive capabilities for novel view synthesis (NVS) tasks. Methods apply it to relighting by separating radiance into BRDF parameters and lighting, yet produce inferior relighting quality with artifacts and unnatural indirect illumination due to the limited capability of each Gaussian, which has constant material parameters and normal, alongside the absence of physical constraints for indirect lighting. In this paper, we present a novel framework called Spatially-vayring Gaussian Inverse Rendering (SVG-IR), aimed at enhancing both NVS and relighting quality. To this end, we propose a new representation—Spatially-varying Gaussian (SVG)—that allows per-Gaussian spatially varying parameters. This enhanced representation is complemented by a SVG splatting scheme akin to vertex/fragment shading in traditional graphics pipelines. Furthermore, we integrate a physically-based indirect lighting model, enabling more realistic relighting. The proposed SVG-IR framework significantly improves rendering quality, outperforming state-of-the-art NeRF-based methods by 2.5 dB in peak signal-to-noise ratio (PSNR) and surpassing existing Gaussian-based techniques by 3.5 dB in relighting tasks, all while maintaining a real-time rendering speed. The source code is available at \href{https://github.com/learner-shx/SVG-IR}{https://github.com/learner-shx/SVG-IR}.

\end{abstract}
\section{Introduction}

Reconstructing 3D assets from images, or so-called \emph{inverse rendering} (IR), is a longstanding task in computer graphics and vision. While recent techniques (e.g., Neural Radiance Fields (NeRF)~\cite{mildenhall_2020_nerf} and 3D Gaussian Splatting (3DGS)~\cite{kerbl3Dgaussians}) have brought great opportunities to this task, it is still challenging due to its ill-posed nature and the complexity of both appearance and lighting.

\begin{figure}[t]
    \centering
    \includegraphics[width=0.95\linewidth]{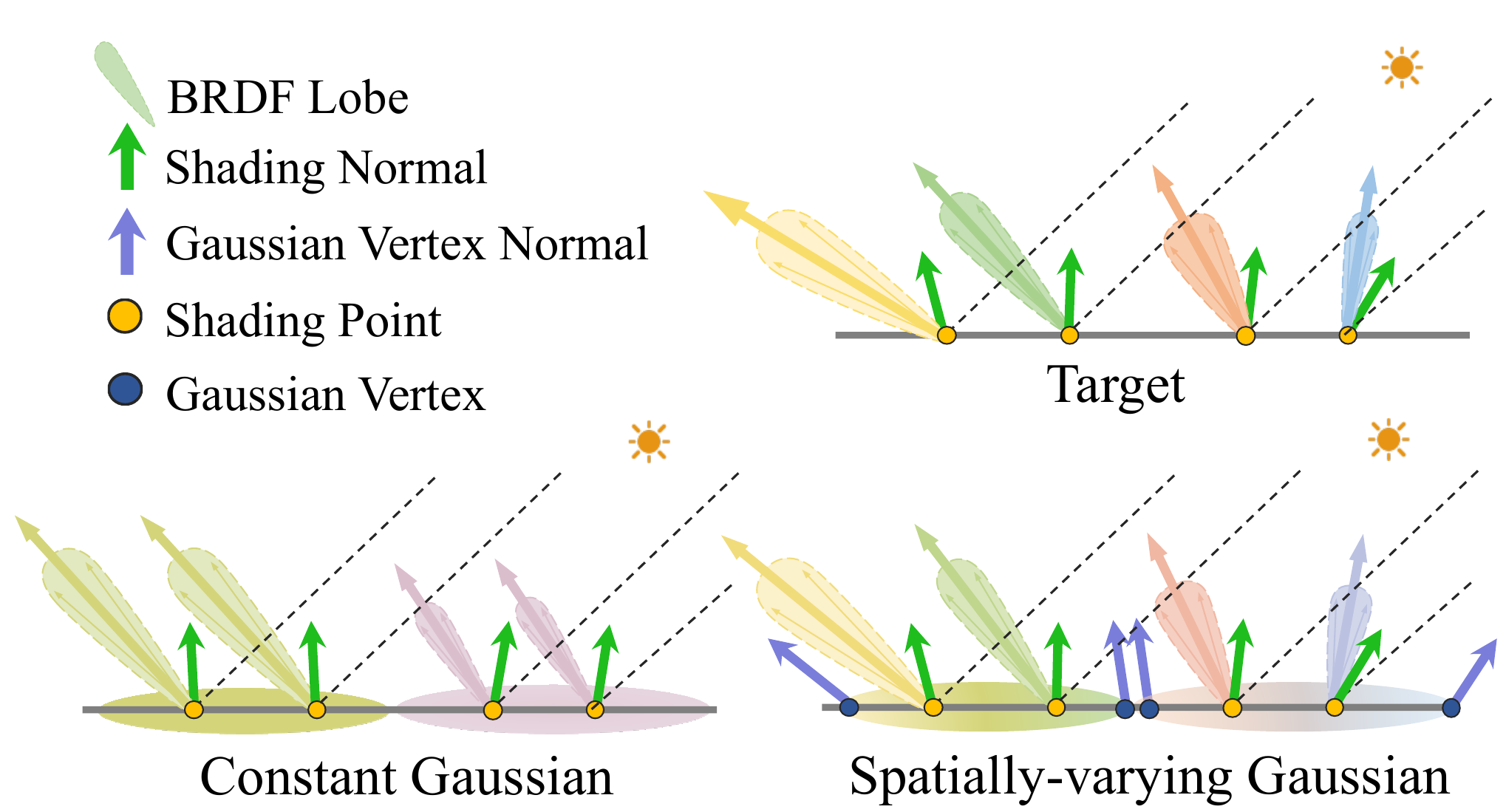}
    \vspace{-6pt}
    \caption{Illustration of two Constant Gaussians and Spatially-varying Gaussians fitting a distribution of BRDFs. As Spatially-varying Gaussians allow spatially-varying parameters, a single Gaussian can have different BRDF lobes at different places, leading to a more flexible representation compared to the Constant Gaussian.}
    \label{fig:curve_gaussian}
    \vspace{-10pt}
\end{figure}

Existing NeRF-based IR methods~\cite{bi_2020_neural,boss_2021_neuralpil,zhang_2021_nerfactor,hasselgren2022shape,zhang_iron_2022,yang_2023_sireir,srinivasa_2021_nerv} have achieved impressive relighting quality, at the cost of long training and rendering time. 
Recently, several methods~\cite{Jiang24GaussianShader,shi_2023_gir,Liang24GS-IR,Gao23R3DG} have incorporated 3DGS into IR by decoupling radiance into material parameters for a Bidirectional Reflectance Distribution Function (BRDF) and environmental lighting to achieve relightability. Despite their fast training and rendering speeds, the quality of relighting often suffers from noticeable artifacts in both training and testing views (as shown in Fig.~\ref{fig:polluted}). The main reason for this issue is that each Gaussian has \emph{constant normal material parameters} and leads to a uniform color across the entire Gaussian for given view and light directions. This constant radiance is against the fact that a Gaussian might cover different pixels that may have varying colors. We illustrate this observation in Fig.~\ref{fig:curve_gaussian}. Furthermore, these works model indirect illumination without considering physical constraints, leading to unnatural indirect lighting, which remains unchanged even under novel lighting conditions.


In this paper, we aim at a more capable representation for appearance modeling and a physically-based model for indirect lighting to enhance both novel view synthesis (NVS) and relighting quality. Inspired by the \emph{flat shading} and \emph{interpolated shading} in the domain of computer graphics, we propose a novel Gaussian representation that generalizes the per-Gaussian constant material to allow spatially-varying material. We refer to the former as the \emph{Constant Gaussian} and the latter as the \emph{Spatially-varying Gaussian} (SVG). The key feature of our Spatially-varying Gaussian is that it allows for different normal and material properties at distinct representative locations, leading to a more powerful representation ability than the Constant Gaussian. Based on the Spatially-varying Gaussian, we design an inverse rendering framework called \emph{SVG-IR}, which consists of two key components. First, we introduce a novel rendering scheme for SVGs, known as SVG splatting, analogous to vertex/fragment shading in computer graphics. Second, we incorporate a physically-based indirect lighting model into our SVG-IR framework by explicitly modeling light transport across different bounces, leading to a more reasonable decoupling of lighting and materials, as well as enabling indirect lighting with novel light sources. As a result, our SVG-IR framework enhances both NVS and relighting quality, outperforming state-of-the-art NeRF-based methods by 2.5 dB in peak signal-to-noise ratio (PSNR) and surpassing Gaussian-based methods by 3.5 dB in the relighting task, maintaining a real-time rendering speed. To summarize, our contributions include: 
\begin{itemize} 
\item a novel representation—Spatially-varying Gaussian—capable of encoding spatially-varying material attributes on single primitive, enhancing representation ability,
 \item a new inverse rendering framework—SVG-IR—built upon the Spatially-varying Gaussian representation, improving both NVS and relighting quality, and
 \item a physically-based indirect lighting model, which explicitly models light transport, resulting in more natural and realistic lighting. 
\end{itemize}

\section{Related Work}

\subsection{Neural Inverse rendering}
Inverse rendering focuses on recovering material, geometry and lighting conditions from multi-view images. Several methods~\cite{bi_2020_neural, zhang_2021_physg, zhang_2022_mii, jin_2023_tensoir, yao2022neilf, zhang_2023_neilfpp} try to decompose lighting and material building upon the geometric formulation and volume rendering with NeRF~\cite{mildenhall_2020_nerf}.
Neural reflectance field~\cite{bi_2020_neural} introduces BRDF and lighting in NeRF. PhySG~\cite{zhang_2021_physg} further refines the lighting model by using spherical Gaussians with incorporated SDFs for more precise geometry reconstruction. Next, several works have focused on modeling indirect illumination and visibility. InvRender~\cite{Zhang22_MII} models indirect illumination by utilizing the neural radiance field. TensoIR~\cite{Jin23_TensoIR} improves indirect lighting efficiency with a compact tri-plane representation for ray tracing. 
NeILF~\cite{yao2022neilf} and NeILF++~\cite{zhang_2023_neilfpp} represent incoming light as a neural incident light field. NeILF++ further combines VolSDF~\cite{yariv_2021_volsdf} with NeILF, incorporating inter-reflections for enhanced performance.

\subsection{Gaussian splatting and relighting}

Recently, 3DGS~\cite{kerbl_2023_3dgs} introduced discretized explicit 3D Gaussian representations for scene modeling, leveraging differentiable rasterization for real-time rendering and achieving impressive results in the NVS task. However, the original 3DGS struggles to produce smooth and natural normals due to the lack of geometric constraints. Existing methods try to resolve this problem by utilizing 2D Gaussians as primitive~\cite{huang_2024_2dgs, Dai2024GaussianSurfels} or modeling a more accurate opacity field~\cite{Yu2024GOF}. Better geometric representations facilitate applications like IR.

IR methods based on Gaussian splatting have emerged by incorporating BRDF into the attributes of Gaussians~\cite{jiang_2024_gaussianshader, liang_2024_gsir, gao_2023_relightablegs,shi_2023_gir}. Considering the illumination modeling, existing methods~\cite{shi_2023_gir, gao_2023_relightablegs, jiang_2024_gaussianshader,liang_2024_gsir} utilize one-step ray tracing or baked-based volumes to obtain visibility and leverage the unconstrained spherical harmonics (SH) to learn residual terms as the indirect illumination, which is challenging to achieve natural indirect lighting, due to the lack of physical constraints. Our approach not only extends the Gaussian representation but also models lighting physically, further improving the decoupling of material and lighting.

\section{Preliminaries}
\mycfiguret{overview}{pipeline_v9.pdf}{Overview of our framework. We propose a novel SVG-IR framework.  Within this framework, we introduce a Spatially-varying Gaussian representation capable of spatially variability with material attributes. We employ SVG splatting, analogous to vertex and fragment shading in the traditional triangle rendering pipeline, to leverage the improved appearance capability of SVGs. Additionally, we present a physics-based lighting model that enforces additional physical constraints to facilitate the decoupling of lighting and material properties.} 
\paragraph{Gaussian surfel splatting.}
Gaussian Surfel Splatting (GSS), also referred to as 2D Gaussian Splatting (2DGS), is a lower-dimensional form of 3DGS. 3DGS uses Gaussians to represent a scene, with the Gaussians defined as
\begin{equation}
    \mathcal{G}(x) = e^{-\frac{1}{2}(x-\mu)^T\Sigma^{-1}(x-\mu)},
\end{equation}
where $\mu$ and $\Sigma$ denotes the mean and covariance matrix of the 3D Gaussian, representing the position and shape of the Gaussian points. $\Sigma$  can be decomposed into the multiplication of a rotation matrix $R$ and a scaling matrix $S$, as $\Sigma=RSS^TR^T$, while the Gaussian Surfels differ as the $S$ is defined in 2-dimensional as,
\begin{equation}
    S = \operatorname{Diag}(s_x, s_y, 0).
    \label{eq:2dscale}
\end{equation}
Similarly to 3DGS, the color $C$ of each pixel is computed by alpha-blending all the ordered Gaussians overlapping the pixel:
\vspace{-10px}
\begin{align}
    T_i &= \prod_{j=1}^{i-1}(1-\alpha_j), \label{eq:trans} \\
    C &= \sum^{n}_{i=0}T_i c_i \alpha_i, \\
    \alpha_i &= \mathcal{G}^{'} (\mathbf{u}; \mathbf{u_i}, \Sigma^{'}_i) o_i.
\end{align}
where $c_i$ is the color of each Gaussian, and $\alpha_i$ is a multiplication between opacity $o_i$ and Gaussian Filter $\mathcal{G}^{'}$~\cite{Zwicker02EWASplatting} by the 2D covariance matrix in the screen space of the Gaussian surfels.

\section{Method}

In this section, we first introduce our new Spatially-varying Gaussian representation (\secref{sec:CG}). Next, we present our inverse rendering framework built on the SVG representation (\secref{sec:CG-IR}), followed by the physically-based illumination model (\secref{sec:pbi}).

\subsection{Spatially-varying Gaussian representation}
\label{sec:CG}

For Gaussian-based inverse rendering approaches, we notice obvious artifacts in both training and testing views, as shown in Fig.~\ref{fig:relight}. The reason behind this phenomenon is that, each Gaussian is tailored with a constant material parameters conducting an uniform color, which is not capable of capturing spatially-varying color across the area of the Gaussian. Therefore, we propose a novel Gaussian representation--\emph{Spatially-varying Gaussian} (SVG), which allows spatially-varying materials, to replace the original Gaussian (i.e., Constant Gaussian). 

In our Spatially-varying Gaussian, we define several representative locations at each Gaussian, where these locations have different material parameters. Analogous to the concept of vertex for a triangle, we name a representative location as a \emph{Gaussian vertex}. The Constant Gaussian is a special case in our definition, which only has one Gaussian vertex. We build the Gaussian vertex based on 2D Gaussian, parameterizing on the tangent space of each Gaussian. The tangent space is defined with rotation matrix $R$ and then is stretched by the scaling matrix $S$. When querying attributes at any locations on the Gaussian, we interpolate the attribute values of Gaussian vertices to achieve spatially-varying properties.

By attaching attributes at Gaussian vertices, the capability of Gaussians can be enhanced to better fit a given distribution, as shown in Fig.~\ref{fig:curve_gaussian}.

\subsection{SVG-IR framework}
\label{sec:CG-IR}

Now, we introduce our inverse rendering framework--\emph{SVG-IR}, on top of the SVG representation. The core of our framework is the rendering of Spatially-varying Gaussians and the physically-based illumination modeling, which will be presented in the next section. 


We use Spatially-varying Gaussians with $M$ Gaussian vertices as our Gaussian primitives, set as 4 in practice. Besides the parameters defined for Gaussians (covariance $\Sigma_i$, position $\mu$, color $c_i$ and opacity $o_i$), we introduce material parameters for each Gaussian vertex: albedo $\boldsymbol{a}_i$ and roughness $r_i$ defined by the Disney Principled BRDF model~\cite{mcauley2013physically}. We also set a normal offset $\Delta{N}_i$ at each Gaussian vertex. Here, the offset indicates the difference from the Gaussian normal ${N}^{g}_i$ defined by Gaussian covariance. Now, Gaussian $i$ has the following attributes:
covariance $\Sigma_i$, position $\mu_i$, radiance color $c_i$, opacity $o_i$, albedo $\boldsymbol{a}^{\{M\}}_i$, roughness $r^{\{M\}}_i$ and normal offset $\Delta{N}^{\{M\}}_i$.

\paragraph{Spatially-varying Gaussian splatting.} Inspired by the rasterization of triangles, we propose a novel rendering scheme for SVG, consisting of \emph{Gaussian vertex shading} and \emph{Gaussian fragment shading}. In the Gaussian vertex shading, the radiance of each Gaussian vertex is computed using the physically-based lighting (see~\secref{sec:pbi}), leading to $M$ colors $\mathbf{c}^{\{M\}_i}$ for each Gaussian. Then, in the fragment shading, given a camera ray, the radiance of each Gaussian is computed by interpolating among $M$ colors, followed by alpha blending, to obtain the radiance of each pixel.

Next, we provide details of Gaussian fragment shading. During rasterization, we compute the coordinates $u,v$ in the tangent space of Gaussian surfel\cite{dai_2024_gaussiansurfels}  $i$ corresponding to pixel $\mathbf{u}$ in the screen space:
\begin{align}
    \mathbf{d}_i &= \mathbf{J}_{pr}^{-1}\left(\mathbf{u}-\mathbf{u}_{i}\right), \\
    (u,v) &= \frac{\mathbf{d}_i}{S[:2] + \delta},
\end{align}
where $\mathbf{d}_i$ is the displacement from the pixel center to the Gaussian center in the tangent space, $\mathbf{J}_{pr}^{-1}$ is the Jacobian of the inverse mapping from a pixel in the image space to the tangent space of each Gaussian and $\delta$ is an offset (set as 0.1).

In practice, we compute pixel's color by bilinear interpolation, followed by alpha blending:
\begin{equation}
\begin{split}
        \mathbf{c}_i &= \text{BilinearInterp}(\mathbf{c}_i^{\{M\}},u,v), \\
        C &= \sum^n_{i=0}{\mathbf{c}_i\alpha_iT_i}.
\end{split}
\end{equation}

\noindent
\textbf{Inverse rendering framework.}
With established Gaussian primitives and their rendering scheme, we build our \hanxiao{multi-stage} inverse rendering framework, as shown in Fig.~\ref{fig:overview}. \hanxiao{Firstly, given multiple views,} we utilize Gaussian Surfels~\cite{dai_2024_gaussiansurfels} to optimize Gaussian attributes as the initialization. Then, we construct Spatially-varying Gaussians with attributes ($\Sigma_i$, $\mu_i$, $c_i$, $o_i$) inherited from the pre-trained Gaussians and the material attributes of Gaussian vertices (i.e., $a_i^{\{M\}}$, $r_i^{\{M\}}$, $\Delta{N}_i^{\{M\}}$). Our framework optimizes the attributes of the Gaussian vertices. Using SVG splatting, we render the images and compute the corresponding loss functions to guide the optimization. 

\begin{figure}[t]
    \centering
    \includegraphics[width=0.95\linewidth]{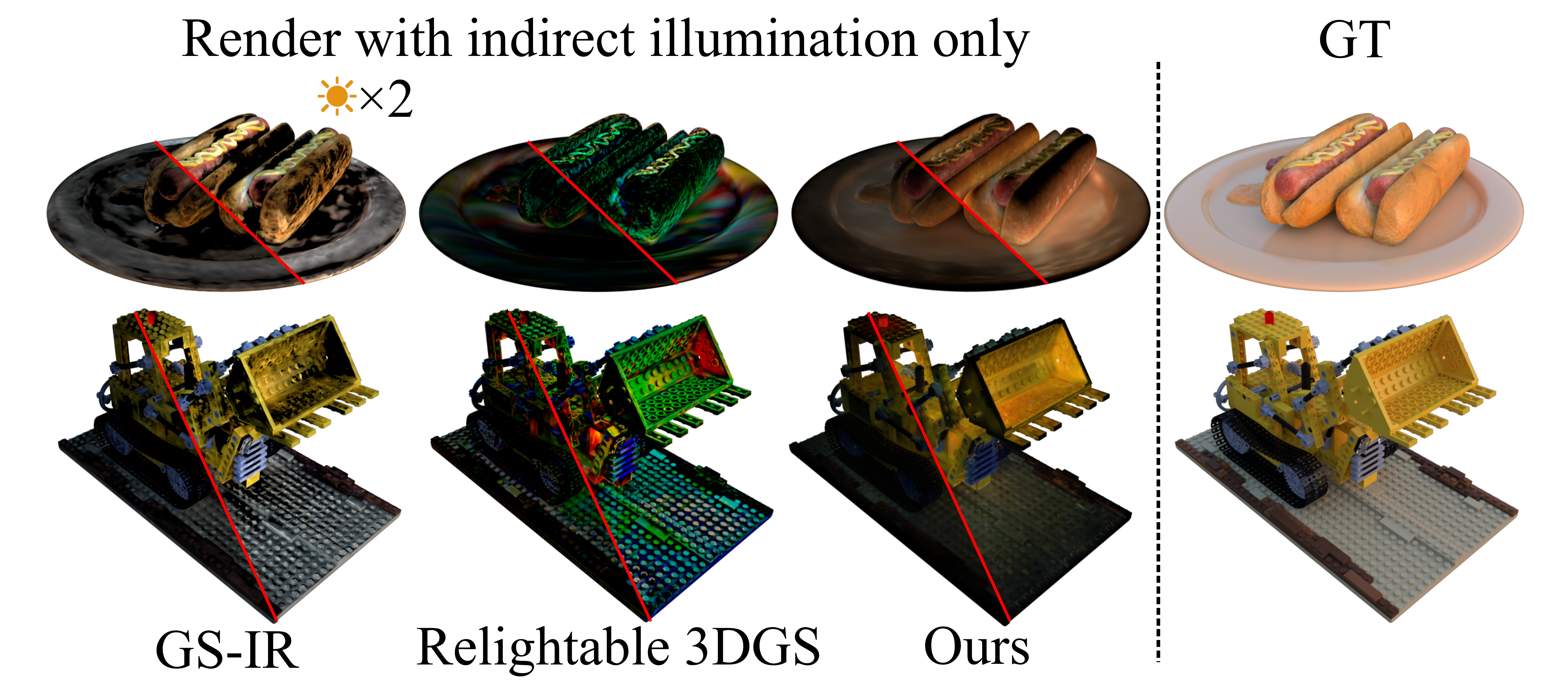}
    \vspace{-6px}
    \caption{The results of rendering using only indirect lighting. We multiplied the brightness of the right half by two for better viewing. GS-IR fails to capture the effects of light reflections from objects as it only models brightness. Meanwhile, Relightable 3DGS lacks supervision for indirect lighting, leading to unnatural results. In contrast, our physically-based lighting model produces more natural rendering results.}
    \label{ii}
    \vspace{-16pt}
\end{figure}

\subsection{Physically-based Illumination}
\label{sec:pbi}
A key of decomposing materials and lighting is the illumination model, including both direct lighting and indirect lighting, particularly for scenes with occlusions. Existing GS-based IR methods model the indirect illumination with trainable SHs which learn either the residual lighting~\cite{Gao23R3DG, shi_2023_gir,liang_2024_gsir} or the residual colors~\cite{Jiang24GaussianShader}. While these unconstrained manner improves NVS quality, the indirect lighting is unnatural, due to the lack of physical constraints. 
We address this issue by introducing physically-based illumination.

In our SVG-IR, we compute the global illumination for each Gaussian vertex in the Gaussian vertex shading. Given a Gaussian vertex $x$, and view direction $\omega_o$, its radiance $L_o(x,\omega_o)$ is defined by the rendering equation~\cite{Kajiya1986RenderingEuqation}: 

\vspace{-10px}
\begin{equation}
      L_o(x, \omega_{o})=\int_{\Omega}f(x,\boldsymbol{a},r,\omega_{i},\omega_{o})L_{i}(x,\omega_{i})(\omega_{i}\cdot n)\mathrm{d}\omega_{i},
      \label{eq:re}
\end{equation}

where $\Omega$ is the upper hemisphere of the Gaussian defined by geometry normal $N^g_i$, $f$ is the Disney Principled BRDF, $\boldsymbol{a}$ is albedo, $r$ is roughness, $\omega_i$ is the incident direction and $n$ is the shading normal of the shading point $x$.

We solve the above integral by uniformly sampling $K$ directions within $\Omega$, and then compute the incoming radiance $L_i(x,\omega_i)$ along each direction. Here, $L_i(x,\omega_i)$ is computed per Gaussian, instead of per Gaussian vertex, as the incoming radiance in a small region tends to be similar. \hanxiao{$L_i(x,\omega_i)$ can be separated as}
\vspace{-6px}
\begin{equation}
       L_i(x, \omega_i) = L_i^{\mathrm{dir}}(x, \omega_i) V(x, \omega_i) + L_i^{\mathrm{ind}}(x, \omega_i),
\label{eq:li}
\end{equation}
\hanxiao{where $L_\mathrm{dir}$ is the direct illumination from the environment map, $V$ is the visibility and $L_\mathrm{ind}$ is our physically-based indirect illumination respectively.}

\paragraph{Physically-based Indirect illumination with raytracing.}

With the pre-trained Gaussian attributes in the initialization, we have the geometry and radiance field (although not relightable). \hanxiao{Besides estimating visibility by ray tracing the same as Relightable 3DGS~\cite{Gao23R3DG}, we compute the indirect incoming radiance $L_i^{\mathrm{ind}}$ by ray tracing in the learned radiance field.}

Specifically, we consider the Gaussians as ellipses, where the axis lengths are scaled by a factor of three. We then construct a bounding volume hierarchy (BVH) over ellipses to accelerate ray tracing. 
For each Gaussian, we uniformly sample $K$ directions on its upper hemisphere. For each sampled direction, we emit a ray starting from the Gaussian center $x$ with direction $\omega_i^k$ and query the Gaussians along the ray using the BVH. Then, we accumulate the transmittance $T$ as~\eqref{eq:trans} and the radiance along each ray as $L_{\mathrm{ind}}$:
\begin{align}
    V(x, \omega_i) &= \begin{cases}
        0, & T \leq \varepsilon, \\
        1, & T > \varepsilon,
    \end{cases}\\
    L_i^{\mathrm{ind}}(x, \omega_i) &= \sum^n_{l=0} c_l \alpha_l T_l, \label{eq:radiance_ind}
\end{align}
where visibility to the environment map $V$ is determined by $T$ and $\varepsilon$ is a threshold, set as 0.8 in practice. $L_{\mathrm{ind}}(x, \omega_i)$ is the indirect radiance accumulated from Gaussian radiance field, $c_l$ and $\alpha_l$ is the radiance field color and transparency of Gaussian $l$. Next, the radiance of Gaussian vertex is computed by~\eqref{eq:re}.

\begin{figure}[t]
    \centering
        \includegraphics[width=\linewidth]{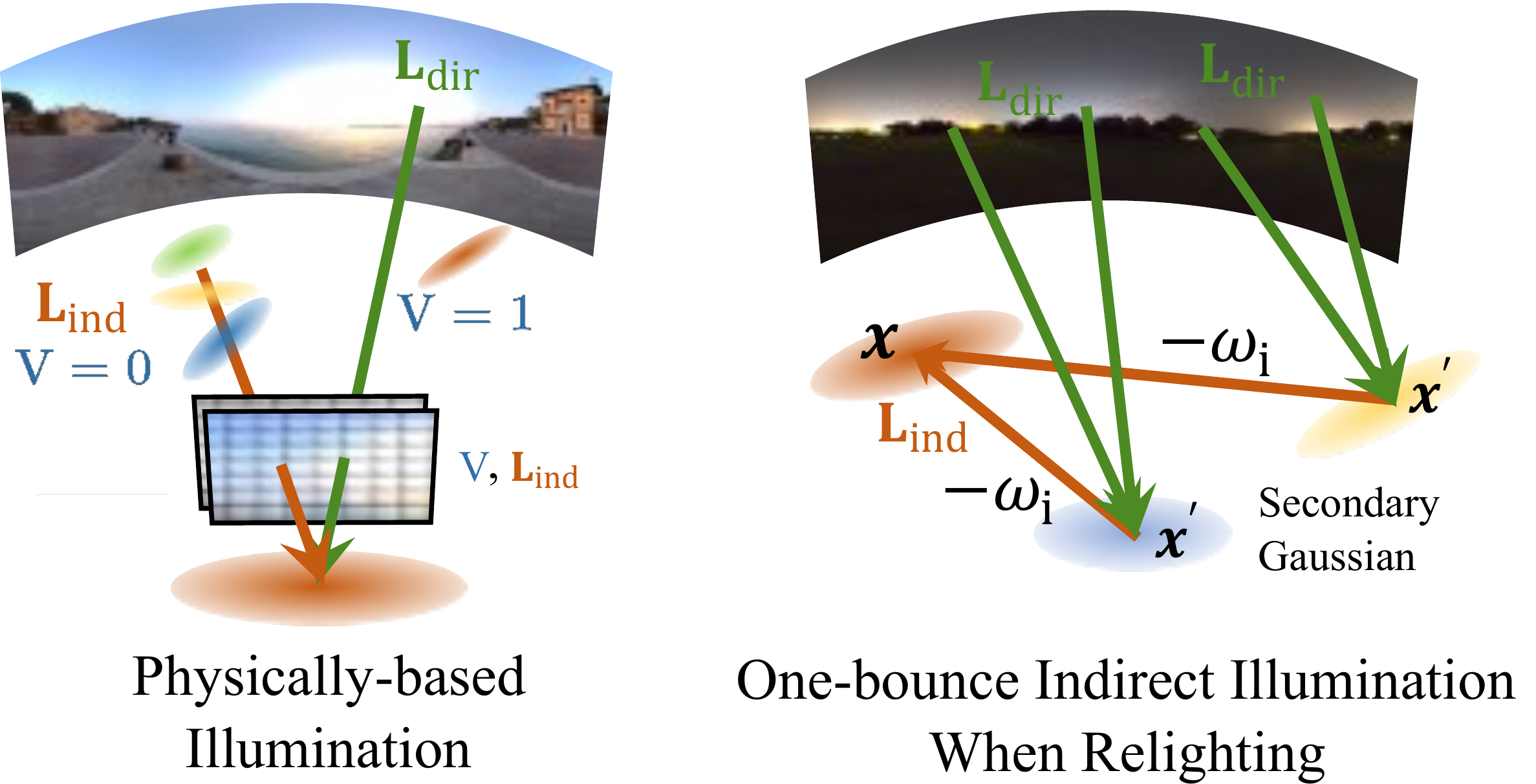}
    \caption{Illustration of the physically-based illumination and our one-bounce relighting method when relighting.}
    \label{fig:pbi}
    \vspace{-16pt}
\end{figure}

\subsection{Relighting with SVG-IR}
\label{sec:relighting}
The learned indirect illumination from previous works~\cite{Gao23R3DG,liang_2024_gsir,Jiang24GaussianShader, shi_2023_gir} is specifically designed for the lighting conditions present during training. As a result, it cannot be used for relighting under new lighting conditions. In this section, we will show how to utilize our trained model for relighting in novel light environments.


\hanxiao{The indirect incoming radiance in~\eqref{eq:radiance_ind} can not be used directly}, as it's learned under the training light. Therefore, we replace $L_i^\mathrm{ind}$ in~\eqref{eq:radiance_ind} with one-bounce indirect illumination for relighting, as shown in Fig.~\ref{fig:pbi}.

For that, we query the Gaussian intersected with the ray at each direction, so called secondary Gaussian, and the incoming radiance $L_i^{\mathrm{ind}}(x,\omega_i)$ at Gaussian vertex $x$ is equivalent to the outgoing radiance $L_o(x',-\omega_i)$ of Gaussian $k$ from $-\omega_i$:
\begin{equation}
    \begin{split}
    &L_i^\mathrm{ind}(x, \omega_i) =  L_o(x', -\omega_i) \\
    & \small =\sum^{K}_{i=0} 2\pi f'(x',\boldsymbol{a}',r',\omega_{i}^k,-\omega_{i})L_{\mathrm{dir}}(x', \omega_{i}^k)V(x', \omega_{i}^k),
    \end{split}
    \label{eq:one-bounce}
    \vspace{-6px}
\end{equation}
where $f'$ represents the interpolated BRDF of vertices on Gaussian $k$. Finally, the radiance of Gaussian vertex is computed by~\eqref{eq:li} and~\eqref{eq:re}.
This allows us to obtain dynamic indirect illumination under the new light source. We show more implementation details in the supplementary material.

\mycfiguret{relight}{results_relight_v3.pdf}{Qualitative comparison on TensoIR Synthetic and ADT datasets. Our method can provide more detailed relighting results than NeRF-based methods. And we alleviate the artifacts of Gaussian-based methods with better decoupling of material and illumination. More results are shown in the supplementary material. }

\definecolor{myyellow}{rgb}{1, 1, 0.7}
\definecolor{myorange}{rgb}{1, 0.85, 0.7}
\definecolor{myred}{rgb}{1, 0.7, 0.7}
\newcommand{\reducedstrut}{\vrule width 0pt height 1.05\ht\strutbox depth 1.0\dp\strutbox\relax}
\newcommand{\sotacolor}[1]{%
  \begingroup
  \setlength{\fboxsep}{0pt}%
  \colorbox{myred}{\reducedstrut#1\/}%
  \endgroup
}
\newcommand{\subsotacolor}[1]{%
  \begingroup
  \setlength{\fboxsep}{0pt}%
  \colorbox{myorange}{\reducedstrut#1\/}%
  \endgroup
}

\begin{table*}[htbp]
  \centering
  \caption{Comparison of relighting performance measured by PSNR$\uparrow$, SSIM$\uparrow$ and LPIPS$\downarrow$ on the TensoIR Synthetic and Aria Digital Twin dataset. Numbers in \sotacolor{red} represent the best performance, while \subsotacolor{orange} numbers denote the second best. Our method not only surpasses existing Gaussian-based approaches but also outperforms previous NeRF-based methods. Furthermore, we maintain the fast training speeds of Gaussian-based methods.}
  
  \resizebox{\textwidth}{!}{ 
  \renewcommand{\arraystretch}{1.5}
   \begin{tabular}{r|l
   @{\hskip 4pt}c@{\hskip 1pt}c@{\hskip 1pt}c@{\hskip 1pt}c@{\hskip 1pt}c
   @{\hskip 4pt}c@{\hskip 1pt}c@{\hskip 1pt}c@{\hskip 1pt}c@{\hskip 1pt}c|
   c@{\hskip 1pt}c@{\hskip 1pt}c@{\hskip 1pt}c@{\hskip 1pt}c
   @{\hskip 4pt}c@{\hskip 1pt}c@{\hskip 1pt}c@{\hskip 1pt}c@{\hskip 1pt}c
   @{\hskip 4pt}c@{\hskip 1pt}c@{\hskip 1pt}c@{\hskip 1pt}c@{\hskip 1pt}c
   @{\hskip 4pt}c@{\hskip 1pt}c@{\hskip 1pt}c@{\hskip 1pt}c@{\hskip 1pt}c}
    \hline
          &       & \multicolumn{10}{c|}{NeRF-based}                                              & \multicolumn{20}{c}{Gaussian-based} \\
          &       & \multicolumn{5}{c}{MII~\cite{zhang_2022_mii}}               & \multicolumn{5}{c|}{TensoIR~\cite{jin_2023_tensoir}}          & \multicolumn{5}{c}{GSshader~\cite{Jiang24GaussianShader}}          & \multicolumn{5}{c}{GS-IR~\cite{liang_2024_gsir}}             & \multicolumn{5}{c}{RelightGS~\cite{gao_2023_relightablegs}}         & \multicolumn{5}{c}{Ours} \\
          &       & PSNR  & /     & SSIM  & /     & LPIPS & PSNR  & /     & SSIM  & /     & LPIPS & PSNR  & /     & SSIM  & /     & LPIPS & PSNR  & /     & SSIM  & /     & LPIPS & PSNR  & /     & SSIM  & /     & LPIPS & PSNR  & /     & SSIM  & /     & LPIPS \\
    \hline
    \multicolumn{1}{c|}{\multirow{5}[4]{*}{\rotatebox{90}{TensoIR}}} & Armadillo & 33.65 & /     & 0.9519 & /     & 0.0643 & \subsotacolor{34.41} & /     & \sotacolor{0.9753} & /     & \sotacolor{0.0449} & 22.76 & /     & 0.9143 & /     & 0.0716 & 29.39 & /     & 0.9204 & /     & 0.0778 & 32.67 & /     & 0.9488 & /     & 0.0697 & \sotacolor{35.01} & /     & \subsotacolor{0.9629} & /     & \subsotacolor{0.0539} \\
          & Ficus & 24.02 & /     & 0.9156 & /     & 0.0758 & 24.30 & /     & \subsotacolor{0.9466} & /     & 0.0680 & 24.16 & /     & 0.9432 & /     & 0.0468 & 25.45 & /     & 0.8934 & /     & 0.0794 & \subsotacolor{29.02} & /     & 0.9399 & /     & \subsotacolor{0.0430} & \sotacolor{31.13} & /     & \sotacolor{0.9543} & /     & \sotacolor{0.0284} \\
          & Hotdog & \subsotacolor{28.91} & /     & 0.9156 & /     & \subsotacolor{0.0945} & 27.88 & /     & \subsotacolor{0.9322} & /     & 0.1160 & 18.02 & /     & 0.8782 & /     & 0.1279 & 21.58 & /     & 0.8899 & /     & 0.1244 & 22.19 & /     & 0.9072 & /     & 0.1127 &\sotacolor{30.02} & /     & \sotacolor{0.9527} & /     & \sotacolor{0.0649} \\
          & Lego  & 24.45 & /     & 0.8513 & /     & 0.1462 &\subsotacolor{27.54} & /     & \sotacolor{0.9249} & /     & \subsotacolor{0.0936} & 14.61 & /     & 0.7920 & /     & 0.1294 & 22.33 & /     & 0.8389 & /     & 0.1141 & 26.52 & /     & 0.8875 & /     & 0.0988 & \sotacolor{28.19} & /     & \subsotacolor{0.9141} & /     & \sotacolor{0.0765} \\
\cline{2-32}          & Mean  & 27.76 & /     & 0.9086 & /     & 0.0952 & \subsotacolor{28.53} & /     & \subsotacolor{0.9448} & /     & \subsotacolor{0.0806} & 19.89 & /     & 0.8819 & /     & 0.0939 & 24.69 & /     & 0.8857 & /     & 0.0989 & 27.60 & /     & 0.9209 & /     & 0.0811 & \sotacolor{31.10} & /     & \sotacolor{0.9460} & /     & \sotacolor{0.0558} \\
    \hline
    \multicolumn{1}{c|}{\multirow{5}[4]{*}{\rotatebox{90}{ADT}}} &Airplane & 31.48 & /     & 0.9770 & /     & \subsotacolor{0.0252} & \subsotacolor{32.63} & /     & \subsotacolor{0.9810} & /     & 0.0324 & 24.81 & /     & 0.9569 & /     & 0.0333 & 28.63 & /     & 0.9558 & /     & 0.0396 & 31.18 & /     & 0.9801 & /     & 0.0283 &\sotacolor{35.98} & /     & \sotacolor{0.9845} & /     & \sotacolor{0.0191} \\
          & Birdhouse & 29.58 & /     & 0.9208 & /     & 0.0837 & \subsotacolor{31.59} & /     & \subsotacolor{0.9551} & /     & 0.0841 & 23.14 & /     & 0.9276 & /     & \subsotacolor{0.0531} & 25.11 & /     & 0.9006 & /     & 0.0813 & 31.05 & /     & 0.9544 & /     & 0.0551 & \sotacolor{31.85} & /     & \sotacolor{0.9592} & /     & \sotacolor{0.0489} \\
          & Gargoyle & 29.16 & /     & 0.9360 & /     & 0.0567 & 30.42 & /     & 0.9609 & /     & 0.0394 & 24.47 & /     & 0.9512 & /     & 0.0356 & 27.78 & /     & 0.9496 & /     & 0.0334 & \subsotacolor{33.97} & /     & \subsotacolor{0.9770} & /     & \subsotacolor{0.0203} & \sotacolor{36.25} & /     &\sotacolor{0.9837} & /     & \sotacolor{0.0146} \\
          & Calculator & 28.73 & /     & 0.9422 & /     & 0.0546 & 27.66 & /     & 0.9549 & /     & 0.0604 & 19.74 & /     & 0.8906 & /     & 0.0818 & 26.26 & /     & 0.9323 & /     & 0.0492 & \subsotacolor{32.07} & /     & \subsotacolor{0.9715} & /     & \subsotacolor{0.0285} & \sotacolor{33.56} & /     & \sotacolor{0.9795} & /     & \sotacolor{0.0226} \\
\cline{2-32}          & Mean  & 29.74 & /     & 0.9440 & /     & 0.0551 & 30.58 & /     & 0.9630 & /     & 0.0541 & 23.04 & /     & 0.9316 & /     & 0.0510 & 26.95 & /     & 0.9346 & /     & 0.0509 & \subsotacolor{32.84} & /     & \subsotacolor{0.9707} & /     & \subsotacolor{0.3180} & \sotacolor{34.69} & /     & \sotacolor{0.9765} & /     & \sotacolor{0.0275} \\
    \hline
          & Training & \multicolumn{5}{c}{6h}                & \multicolumn{5}{c|}{5h}               & \multicolumn{5}{c}{0.5h}              & \multicolumn{5}{c}{0.5h}              & \multicolumn{5}{c}{1h}                & \multicolumn{5}{c}{1h} \\
    \hline
    \end{tabular}%
  }
  \vspace{-8pt}
  \label{tab:relight}%
\end{table*}%

\begin{table}[htbp]
  \centering
  \caption{Comparison of NVS quality between our method and others on the TensoIR Synthetic and ADT dataset. \sotacolor{Red} numbers represent the best performance, and \subsotacolor{orange} denotes the second best.}
  
  \resizebox{0.8 \linewidth}{!}{ 
  \renewcommand{\arraystretch}{1.35}
  
   \begin{tabular}{l
   @{\hskip 4pt}c@{\hskip 1pt}c@{\hskip 1pt}c@{\hskip 1pt}c@{\hskip 1pt}c
   @{\hskip 4pt}c@{\hskip 1pt}c@{\hskip 1pt}c@{\hskip 1pt}c@{\hskip 1pt}c
   @{\hskip 4pt}c@{\hskip 1pt}c@{\hskip 1pt}c@{\hskip 1pt}c@{\hskip 1pt}c}
    \toprule
          & \multicolumn{5}{c}{TensoIR}           & \multicolumn{5}{c}{ADT} \\
          & PSNR  & /     & SSIM  & /     & LPIPS & PSNR  & /     & SSIM  & /     & LPIPS \\
    \midrule
    MII   & 30.92 & /     & 0.9371 & /     & 0.0801 & 30.94 & /     & 0.9568 & /     & 0.0506 \\
    TensoIR & \subsotacolor{35.17} & /     & \subsotacolor{0.9764} & /     & \subsotacolor{0.0397} & \subsotacolor{39.72} & /     & \subsotacolor{0.9916} & /     & \subsotacolor{0.0173} \\
    GS-IR & 35.02 & /     & 0.9637 & /     & 0.0429 & 37.30 & /     & 0.9821 & /     & 0.0219 \\
    RelightGS & 33.35 & /     & 0.9657 & /     & 0.0414 & 36.94 & /     & 0.9879 & /     & 0.0146 \\
    \midrule
    Ours  & \sotacolor{36.71} & /     & \sotacolor{0.9758} & /     & \sotacolor{0.0332} &\sotacolor{41.48} & /     & \sotacolor{0.9923} & /     & \sotacolor{0.0103} \\
    \bottomrule
    \end{tabular}%
  }
  \vspace{-16pt}
  \label{tab:NVS}%
\end{table}%

\section{Implementation details}

\subsection{Radiance consistency loss}
\label{sec:rc_loss}
We observe that the indirect incoming radiance, obtained through ray tracing in the radiance field, can serve as a supervision for the outgoing radiance of secondary Gaussians along the sampled $K$ directions. This provides additional guidance from extra viewpoints. However, performing one-bounce ray tracing for all Gaussians along the $K$ directions is costly. We only sample one direction near the specular reflection for each Gaussian $j$. The supervision is applied as follows:
\begin{equation}
    \mathcal{L}_{\mathrm{rc}} = |L_{i}^{\mathrm{ind}}(x_j,\omega_i^k), L_{i}^{\mathrm{ind}'}(x_j,\omega_i^k)|, 
    \label{eq:rad_loss}
\end{equation}
where $L_{i}^{\mathrm{ind}'}$ is the outgoing radiance of the secondary Gaussian on direction $\omega_i^k$. For more details, please refer to the supplementary material.

\subsection{Training details}
\label{sec:trainingdetails}
We train our model using Adam optimizer~\cite{kingma_2014_adam} by the utilizing the losses as
\begin{equation}
\begin{split}
    \mathcal{L}=&\lambda_{1} \mathcal{L}_{1}+\lambda_{\text{ssim}} \mathcal{L}_{\text{ssim}}+\lambda_{\text{rc}} \mathcal{L}_{\text{rc}} \\
    &\lambda_{n} \mathcal{L}_{n}+\lambda_{s, a} \mathcal{L}_{s, a}+\lambda_{s, r} \mathcal{L}_{s, r}+\lambda_{\mathrm{reg},n} \mathcal{L}_{\mathrm{reg},n},
    \end{split}
    \label{eq:loss}
\end{equation}
    where $\mathcal{L}_{1}$ and $\mathcal{L}_{\text{ssim}}$ is $L_1$ loss and SSIM loss between rendered image and ground truth, $\mathcal{L}_{\text{rc}}$ is the radiance consistency loss in Sec.~\ref{sec:rc_loss}, $\mathcal{L}_{n}$ is one minus cosine similarity between normal and pseudo normal, $\mathcal{L}_{s, a}$ and $\mathcal{L}_{s, r}$ are TV-loss on albedo and roughness, $\mathcal{L}_{\mathrm{reg},n}$ is $L_2$ regular term of normal offsets $\Delta N^M$. More details are in the supplementary. \

\begin{figure*}[t]
    \centering
    \includegraphics[width=0.95\textwidth]{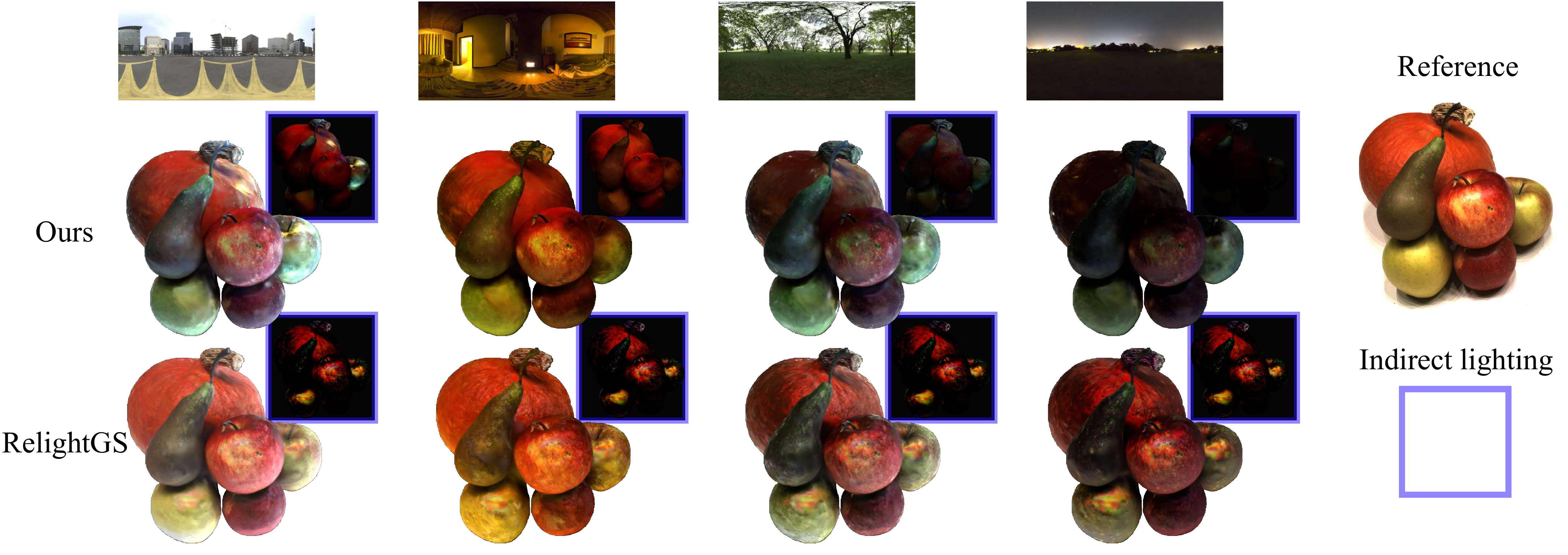}
    \caption{\textbf{Relighting results.} We perform relighting on real dataset DTU. The indirect shading results are in the blue borders. By recalculating indirect lighting for each new light source, our method avoids the unrealistic bright indirect illumination that Relightable 3DGS~\cite{Gao23R3DG} exhibits under low-light conditions. More results are in the supplementary material.}
    \label{fig:religt_dtu}
    \vspace{-14pt}
\end{figure*}
 \vspace{-10pt}

\section{Results}

\label{sec:results}

\subsection{Evaluation setup}
\label{sec:eval_setup}

\paragraph{Dataset.} To evaluate the capabilities on NVS and relighting of our method, we conduct experiments on both synthetic and real-world datasets. For synthetic datasets, we benchmark on the commonly used TensoIR Synthetic datasets~\cite{jin_2023_tensoir} and our collected relighting datasets of four real world scanned objects from Aria Digital Twin (ADT)~\cite{pan2023aria}. This dataset will be released. And the real-world datasets include DTU~\cite{jensen2014large} and MipNeRF360~\cite{barron_2022_mipnerf360}. Results on more datasets including NeRF Synthetic~\cite{mildenhall_2020_nerf} and NeILF++~\cite{zhang_2023_neilfpp} are in the supplementary material. 

\noindent
\textbf{Methods for comparison.} We selected representative NeRF-based inverse rendering methods MII~\cite{zhang_2022_mii}, TensoIR~\cite{jin_2023_tensoir}, as well as recent GS-based methods, such as GS-IR~\cite{liang_2024_gsir}, GaussianShader~\cite{Jiang24GaussianShader} and Relightable 3DGS~\cite{gao_2023_relightablegs}. We trained each method using the official code and settings.
\paragraph{Metrics.} We evaluate the quality of our comparison and ablation study results using PSNR, structural similarity index (SSIM)~\cite{wang_2004_ssim}, and learned perceptual image patch similarity (LPIPS)~\cite{zhang_2018_lpips}. To account for the ambiguity between albedo and lighting, we follow TensoIR by rescaling relighting images according to both the ground truth and predicted albedo. Additionally, we report the mean training time, both tested on the same one RTX 4090 setup.

\subsection{Comparison with previous works}
\label{sec:res}

\paragraph{Relighting on synthetic data.} In Tab.~\ref{tab:relight}, we provide the numerical relighting results on the TensoIR Synthetic and ADT datasets. Our method achieves the SOTA relighting results in almost all scenes. In Fig.~\ref{fig:relight}, we provide the visual comparison between our method and others. Our method produces smoother and more detailed relighting renderings, thanks to our Spatially-varying Gaussian representation and physically-based illumination. In contrast, GS-IR shows noticeable artifacts due to the constant color on a single Gaussian, and TensoIR produces renderings with missing details and unnatural lighting effects. Additional metrics on the datasets, along with more images, are provided in the supplementary materials.

\begin{figure}[t]
    \centering
    \includegraphics[width=\linewidth]{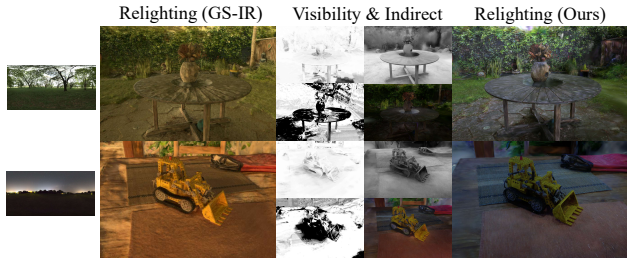}
    \caption{\textbf{Indirect Illumination on real dataset MipNeRF360.} Our physically-based indirect illumination modeling helps us achieve natural relighting results. More results are in the supplementary material.}
    \label{fig:result_ii}
   \vspace{-14pt}
\end{figure}

\paragraph{Relighting on real data.}
For real scenes, we conducted experiments on several scenes selected from the object-level datasets DTU datasets and scene-level dataset MipNeRF360. Since ground truth is not available, we provide the reference training views along with the rendering results under novel lighting conditions. As shown in Fig.~\ref{fig:religt_dtu}, the indirect lighting renderings produced by Relightable 3DGS under different environment maps remain the same, causing unnaturally bright renderings even under dark light sources. On the contrary, our method can dynamically adopt indirect lighting w.r.t. different environment maps due to our relighting strategy, which uses one-bounce ray tracing to model indirect illumination. Similarly, we compare our method with GS-IR in Fig.~\ref{fig:result_ii}, showing more natural relighting results. 

\begin{figure}[t]
    \centering
    \includegraphics[width=\linewidth]{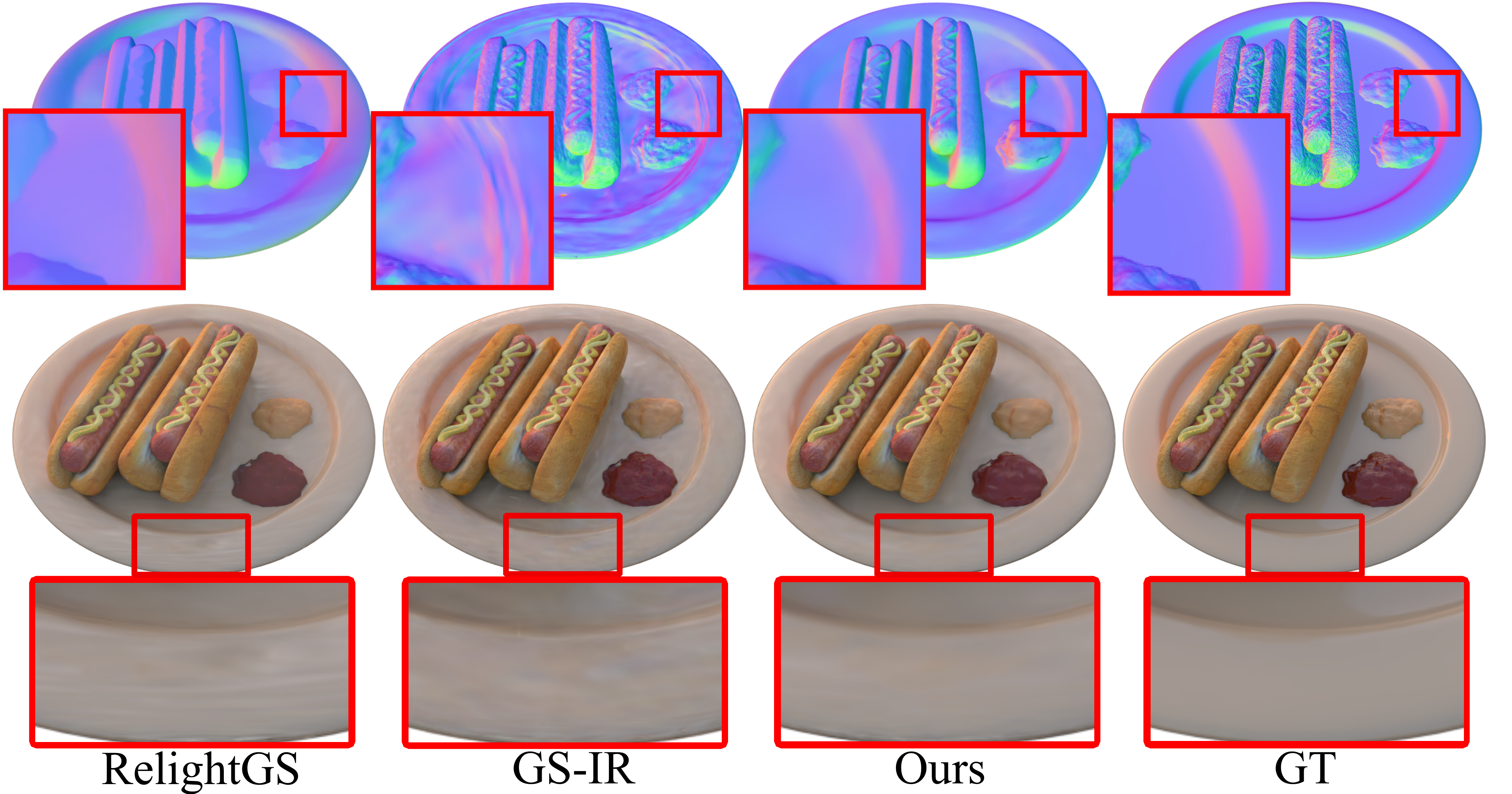}
    \vspace{-18pt}
    \caption{Comparison between our method and other GS-based IR methods on the normal reconstruction and NVS results. Our Spatially-varying Gaussian representation enhances the representation capacity, producing less artifacts.}
   \vspace{-8pt}
    \label{fig:polluted}
\end{figure}

\paragraph{NVS.}
Besides relighting task, we validate our method on the NVS task, by performing a comparison with other IR methods~\cite{Zhang22_MII,jin_2023_tensoir,liang_2024_gsir,gao_2023_relightablegs}. In Tab.~\ref{tab:NVS}, our method outperforms the others on the TensoIR Synthetic and ADT datasets. The visual comparison is shown in Fig.~\ref{fig:polluted}, showing that our method can produce cleaner results than GS-IR~\cite{liang_2024_gsir} and Reglitable 3DGS~\cite{gao_2023_relightablegs} with less artifacts, thanks to the spatially-varying material in a single Gaussian. 

\subsection{Ablation study}
\label{sec:ab}
We perform ablation studies on the key components of our model on the TensoIR Synthetic dataset in terms of PSNR, SSIM and LPIPS. As shown in Tab.~\ref{tab:ab_key}, we observe that as we incorporate our key components, the performance improves consistently.

\begin{figure}[t]
    \centering
    \includegraphics[width=\linewidth]{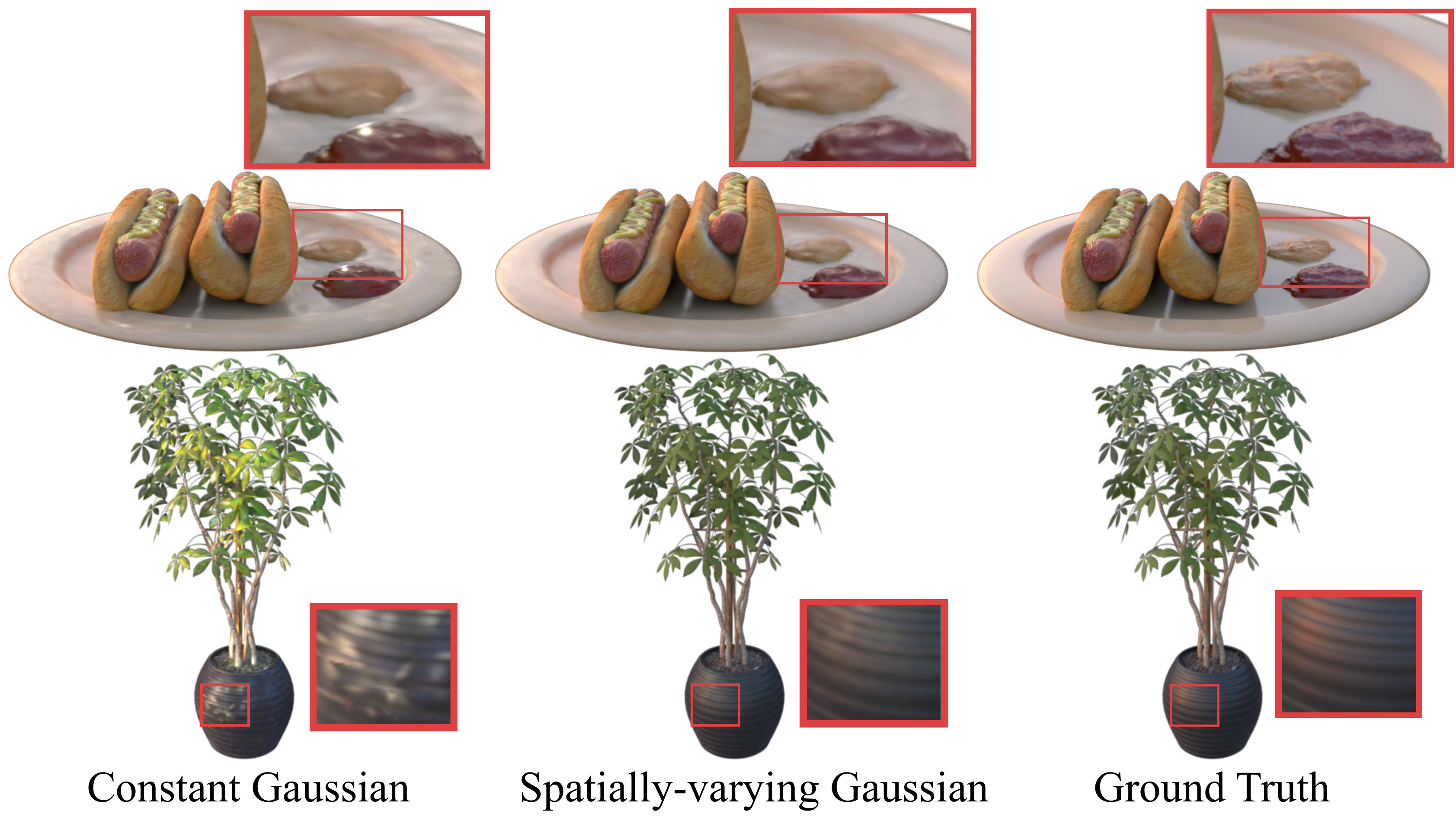}
    \vspace{-10pt}
    \caption{Ablation study on Spatially-varying Gaussian representation. The Spatially-varying Gaussian yields smoother rendering results, significantly reducing artifacts.}
   \vspace{-10pt}
    \label{fig:ab_curve}
\end{figure}

\paragraph{Spatially-varying Gaussian representation.}
We evaluate the impact of our Spatially-varying Gaussian representation in Tab.~\ref{tab:ab_key} and Fig.~\ref{fig:ab_curve}. As shown in Fig~\ref{fig:ab_curve}, our Spatially-varying Gaussian shows a smoother surface, and fewer artifacts compared to the Constant Gaussian.

\begin{table}[htbp]
  \centering
  \caption{Ablation study of our key components on TensoIR Synthetic dataset. ``SVG'' means the Spatially-varying Gaussian representation, ``PBI'' means the physically-based illumination. Numbers in \sotacolor{red} represent the best performance, while \subsotacolor{orange} numbers denote the second best.}
  \resizebox{\linewidth}{!}{
    \begin{tabular}{rrcccccc}
    \toprule
    \multicolumn{2}{c}{Component} & \multicolumn{3}{c}{Relighting} & \multicolumn{3}{c}{NVS} \\
    \midrule
    \multicolumn{1}{c}{SVG} & \multicolumn{1}{c}{PBI} & PSNR↑ & SSIM↑ & LPIPS↓ & PSNR↑ & SSIM↑ & LPIPS↓ \\
    \midrule
    \ding{55}      &   \ding{55}    & 28.614 & 0.904 & 0.093 & 34.640 & 0.941 & 0.047 \\
    \ding{51}     &  \ding{55}     & \subsotacolor{29.447} & \subsotacolor{0.937} & \subsotacolor{0.074} & \subsotacolor{35.794} & \subsotacolor{0.961} & \subsotacolor{0.041} \\
    \ding{51}     & \ding{51}     & \sotacolor{31.087} & \sotacolor{0.946} & \sotacolor{0.055} & \sotacolor{36.709} & \sotacolor{0.975} & \sotacolor{0.033} \\
    \bottomrule
    \end{tabular}%
    }
  \label{tab:ab_key}%
  \vspace{-8pt}
\end{table}%

\noindent
\textbf{Indirect illumination modeling.}
Besides the metrics improvement shown in Tab.~\ref{tab:ab_key}, 
we further conducted a more in-depth ablation study on components (visibility, indirect illumination) of our physically-based indirect illumination, as shown in Tab.~\ref{tab:ab_ii}. As the components were gradually added, both the relighting and NVS quality progressively improved. The specific impact of each component on the training process is explained in the supplementary.



\begin{table}[htbp]
  \centering
  \caption{Ablation study of several components in our physically-based lighting model on the TensoIR Synthetic dataset. ``Vis.'' means the visibility to environment light, ``Ind.'' means the radiance field indirect illumination term. Numbers in \sotacolor{red} represent the best performance, while \subsotacolor{orange} numbers denote the second best.}
  \resizebox{\linewidth}{!}{
  \setlength{\tabcolsep}{4pt} 
    \begin{tabular}{rrcccccc}
    \toprule
    \multicolumn{2}{c}{Component} & \multicolumn{3}{c}{Relighting} & \multicolumn{3}{c}{NVS} \\
    \midrule
    \multicolumn{1}{l}{Vis.} & \multicolumn{1}{l}{Ind.} & \multicolumn{1}{l}{PSNR↑} & \multicolumn{1}{l}{SSIM↑} & \multicolumn{1}{l}{LPIPS↓} & \multicolumn{1}{l}{PSNR↑} & \multicolumn{1}{l}{SSIM↑} & \multicolumn{1}{l}{LPIPS↓} \\
    \midrule
    \ding{55}      &\ding{55}          & 29.447 & 0.937 & 0.074 & 35.794 & 0.961 & 0.041 \\
    \ding{51}     & \ding{55}          & \subsotacolor{30.253} & \subsotacolor{0.941} & \subsotacolor{0.068} & \subsotacolor{36.059} & \subsotacolor{0.963} & \subsotacolor{0.037} \\
    \ding{51}     & \ding{51}         & \sotacolor{31.087} & \sotacolor{0.946} & \sotacolor{0.055} & \sotacolor{36.709} & \sotacolor{0.975} & \sotacolor{0.033} \\
    \bottomrule
    \end{tabular}%
    }
  \label{tab:ab_ii}%
\end{table}%

\subsection{Discussion and limitations}
\label{sec:lim}
We have identified several limitations of our method. We do not introduce geometric prior, leading to unsatisfied recovery for highly specular objects. This issue can be alleviated by introducing SDF like GS-ROR~\cite{zhu2024gs-ror}, which we leave for future work. \hanxiao{Additionally, our model brings 50\% $\sim$ 80\% more GPU memory cost compared to 3DGS/2DGS and 10\% $\sim$ 20\% more compared to Relightable 3DGS~\cite{Gao23R3DG}, due to the Gaussian vertices.}


\section{Conclusion}

In this paper, we have presented a novel framework for inverse rendering. At the core of our framework is a spatially-varying Gaussian representation and a curve Gaussian rendering scheme, which allows spatially-varying material and normal parameters, showing more powerful representation capability than typical Gaussian primitives. Besides, we present a physically-based indirect illumination modeling method, which brings physical constraints for training and enables natural indirect light for relighting. Consequently, our method outperforms existing NeRF-based and Gaussian-based inverse rendering methods without losing efficiency. In future work, improving the geometric accuracy is a potential direction.

  

\section*{Acknowledgment}

We thank the reviewers for the valuable comments. This work has been partially supported by the National Science and Technology Major Project under grant No. 2022ZD0116305 and National Natural Science Foundation of China under grant No. 62172220.

{
    \small
    \bibliographystyle{ieeenat_fullname}
    \bibliography{paper}

\begin{thebibliography}{39}
\providecommand{\natexlab}[1]{#1}
\providecommand{\url}[1]{\texttt{#1}}
\expandafter\ifx\csname urlstyle\endcsname\relax
  \providecommand{\doi}[1]{doi: #1}\else
  \providecommand{\doi}{doi: \begingroup \urlstyle{rm}\Url}\fi

\bibitem[Barron et~al.(2022)Barron, Mildenhall, Verbin, Srinivasan, and Hedman]{barron_2022_mipnerf360}
Jonathan~T. Barron, Ben Mildenhall, Dor Verbin, Pratul~P. Srinivasan, and Peter Hedman.
\newblock Mip-nerf 360: Unbounded anti-aliased neural radiance fields.
\newblock In \emph{CVPR}, pages 5470--5479, New York, NY, USA, 2022. IEEE.

\bibitem[Bi et~al.(2020)Bi, Xu, Srinivasan, Mildenhall, Sunkavalli, Hašan, Hold-Geoffroy, Kriegman, and Ramamoorthi]{bi_2020_neural}
Sai Bi, Zexiang Xu, Pratul Srinivasan, Ben Mildenhall, Kalyan Sunkavalli, Miloš Hašan, Yannick Hold-Geoffroy, David Kriegman, and Ravi Ramamoorthi.
\newblock Neural reflectance fields for appearance acquisition, 2020.

\bibitem[Boss et~al.(2021)Boss, Jampani, Braun, Liu, Barron, and Lensch]{boss_2021_neuralpil}
Mark Boss, Varun Jampani, Raphael Braun, Ce Liu, Jonathan Barron, and Hendrik~PA Lensch.
\newblock Neural-pil: Neural pre-integrated lighting for reflectance decomposition.
\newblock In \emph{NeurIPS}, pages 10691--10704, Red Hook, NY, USA, 2021. Curran Associates, Inc.

\bibitem[Dai et~al.(2024{\natexlab{a}})Dai, Xu, Xie, Liu, Wang, and Xu]{Dai2024GaussianSurfels}
Pinxuan Dai, Jiamin Xu, Wenxiang Xie, Xinguo Liu, Huamin Wang, and Weiwei Xu.
\newblock High-quality surface reconstruction using gaussian surfels.
\newblock In \emph{ACM SIGGRAPH 2024 Conference Papers}. Association for Computing Machinery, 2024{\natexlab{a}}.

\bibitem[Dai et~al.(2024{\natexlab{b}})Dai, Xu, Xie, Liu, Wang, and Xu]{dai_2024_gaussiansurfels}
Pinxuan Dai, Jiamin Xu, Wenxiang Xie, Xinguo Liu, Huamin Wang, and Weiwei Xu.
\newblock High-quality surface reconstruction using gaussian surfels.
\newblock In \emph{SIGGRAPH}, New York, NY, USA, 2024{\natexlab{b}}. Association for Computing Machinery.

\bibitem[Gao et~al.(2023)Gao, Gu, Lin, Zhu, Cao, Zhang, and Yao]{Gao23R3DG}
Jian Gao, Chun Gu, Youtian Lin, Hao Zhu, Xun Cao, Li Zhang, and Yao Yao.
\newblock Relightable 3d gaussian: Real-time point cloud relighting with brdf decomposition and ray tracing.
\newblock \emph{arXiv:2311.16043}, 2023.

\bibitem[Gao et~al.(2024)Gao, Gu, Lin, Li, Zhu, Cao, Zhang, and Yao]{gao_2023_relightablegs}
Jian Gao, Chun Gu, Youtian Lin, Zhihao Li, Hao Zhu, Xun Cao, Li Zhang, and Yao Yao.
\newblock Relightable 3d gaussians: Realistic point cloud relighting with brdf decomposition and ray tracing, 2024.

\bibitem[Hasselgren et~al.(2022)Hasselgren, Hofmann, and Munkberg]{hasselgren2022shape}
Jon Hasselgren, Nikolai Hofmann, and Jacob Munkberg.
\newblock Shape, light, and material decomposition from images using monte carlo rendering and denoising.
\newblock \emph{Advances in Neural Information Processing Systems}, 35:\penalty0 22856--22869, 2022.

\bibitem[Huang et~al.(2024)Huang, Yu, Chen, Geiger, and Gao]{huang_2024_2dgs}
Binbin Huang, Zehao Yu, Anpei Chen, Andreas Geiger, and Shenghua Gao.
\newblock 2d gaussian splatting for geometrically accurate radiance fields.
\newblock In \emph{SIGGRAPH}, New York, NY, USA, 2024. Association for Computing Machinery.

\bibitem[Jensen et~al.(2014)Jensen, Dahl, Vogiatzis, Tola, and Aan{\ae}s]{jensen2014large}
Rasmus Jensen, Anders Dahl, George Vogiatzis, Engil Tola, and Henrik Aan{\ae}s.
\newblock Large scale multi-view stereopsis evaluation.
\newblock In \emph{2014 IEEE Conference on Computer Vision and Pattern Recognition}, pages 406--413. IEEE, 2014.

\bibitem[Jiang et~al.(2024{\natexlab{a}})Jiang, Tu, Liu, Gao, Long, Wang, and Ma]{Jiang24GaussianShader}
Yingwenqi Jiang, Jiadong Tu, Yuan Liu, Xifeng Gao, Xiaoxiao Long, Wenping Wang, and Yuexin Ma.
\newblock Gaussianshader: 3d gaussian splatting with shading functions for reflective surfaces.
\newblock In \emph{2024 IEEE/CVF Conference on Computer Vision and Pattern Recognition (CVPR)}, pages 5322--5332, 2024{\natexlab{a}}.

\bibitem[Jiang et~al.(2024{\natexlab{b}})Jiang, Tu, Liu, Gao, Long, Wang, and Ma]{jiang_2024_gaussianshader}
Yingwenqi Jiang, Jiadong Tu, Yuan Liu, Xifeng Gao, Xiaoxiao Long, Wenping Wang, and Yuexin Ma.
\newblock Gaussianshader: 3d gaussian splatting with shading functions for reflective surfaces.
\newblock In \emph{CVPR}, pages 5322--5332, New York, NY, USA, 2024{\natexlab{b}}. IEEE.

\bibitem[Jin et~al.(2023{\natexlab{a}})Jin, Liu, Xu, Zhang, Han, Bi, Zhou, Xu, and Su]{Jin23_TensoIR}
Haian Jin, Isabella Liu, Peijia Xu, Xiaoshuai Zhang, Songfang Han, Sai Bi, Xiaowei Zhou, Zexiang Xu, and Hao Su.
\newblock Tensoir: Tensorial inverse rendering.
\newblock In \emph{2023 IEEE/CVF Conference on Computer Vision and Pattern Recognition (CVPR)}, pages 165--174, 2023{\natexlab{a}}.

\bibitem[Jin et~al.(2023{\natexlab{b}})Jin, Liu, Xu, Zhang, Han, Bi, Zhou, Xu, and Su]{jin_2023_tensoir}
Haian Jin, Isabella Liu, Peijia Xu, Xiaoshuai Zhang, Songfang Han, Sai Bi, Xiaowei Zhou, Zexiang Xu, and Hao Su.
\newblock Tensoir: Tensorial inverse rendering.
\newblock In \emph{CVPR}, pages 165--174, New York, NY, USA, 2023{\natexlab{b}}. IEEE.

\bibitem[Kajiya(1986)]{Kajiya1986RenderingEuqation}
James~T. Kajiya.
\newblock The rendering equation.
\newblock \emph{SIGGRAPH Comput. Graph.}, 20\penalty0 (4):\penalty0 143–150, 1986.

\bibitem[Kerbl et~al.(2023{\natexlab{a}})Kerbl, Kopanas, Leimk{\"u}hler, and Drettakis]{kerbl3Dgaussians}
Bernhard Kerbl, Georgios Kopanas, Thomas Leimk{\"u}hler, and George Drettakis.
\newblock 3d gaussian splatting for real-time radiance field rendering.
\newblock \emph{ACM Transactions on Graphics}, 42\penalty0 (4), 2023{\natexlab{a}}.

\bibitem[Kerbl et~al.(2023{\natexlab{b}})Kerbl, Kopanas, Leimk{\"u}hler, and Drettakis]{kerbl_2023_3dgs}
Bernhard Kerbl, Georgios Kopanas, Thomas Leimk{\"u}hler, and George Drettakis.
\newblock 3d gaussian splatting for real-time radiance field rendering.
\newblock \emph{ACM TOG}, 42\penalty0 (4):\penalty0 139--1, 2023{\natexlab{b}}.

\bibitem[Kingma and Ba(2017)]{kingma_2014_adam}
Diederik~P. Kingma and Jimmy Ba.
\newblock Adam: A method for stochastic optimization, 2017.

\bibitem[Liang et~al.(2024{\natexlab{a}})Liang, Zhang, Feng, Shan, and Jia]{Liang24GS-IR}
Zhihao Liang, Qi Zhang, Ying Feng, Ying Shan, and Kui Jia.
\newblock Gs-ir: 3d gaussian splatting for inverse rendering.
\newblock In \emph{2024 IEEE/CVF Conference on Computer Vision and Pattern Recognition (CVPR)}, pages 21644--21653, 2024{\natexlab{a}}.

\bibitem[Liang et~al.(2024{\natexlab{b}})Liang, Zhang, Feng, Shan, and Jia]{liang_2024_gsir}
Zhihao Liang, Qi Zhang, Ying Feng, Ying Shan, and Kui Jia.
\newblock Gs-ir: 3d gaussian splatting for inverse rendering.
\newblock In \emph{CVPR}, pages 21644--21653, New York, NY, USA, 2024{\natexlab{b}}. IEEE.

\bibitem[McAuley et~al.(2013)McAuley, Hill, Martinez, Villemin, Pettineo, Lazarov, Neubelt, Karis, Hery, Hoffman, et~al.]{mcauley2013physically}
Stephen McAuley, Stephen Hill, Adam Martinez, Ryusuke Villemin, Matt Pettineo, Dimitar Lazarov, David Neubelt, Brian Karis, Christophe Hery, Naty Hoffman, et~al.
\newblock Physically based shading in theory and practice.
\newblock In \emph{ACM SIGGRAPH 2013 Courses}, pages 1--8. 2013.

\bibitem[Mildenhall et~al.(2020)Mildenhall, Srinivasan, Tancik, Barron, Ramamoorthi, and Ng]{mildenhall_2020_nerf}
Ben Mildenhall, Pratul~P. Srinivasan, Matthew Tancik, Jonathan~T. Barron, Ravi Ramamoorthi, and Ren Ng.
\newblock Nerf: Representing scenes as neural radiance fields for view synthesis.
\newblock In \emph{ECCV}, pages 405--421, Berlin, Heidelberg, 2020. Springer.

\bibitem[Pan et~al.(2023)Pan, Charron, Yang, Peters, Whelan, Kong, Parkhi, Newcombe, and Ren]{pan2023aria}
Xiaqing Pan, Nicholas Charron, Yongqian Yang, Scott Peters, Thomas Whelan, Chen Kong, Omkar Parkhi, Richard Newcombe, and Carl~Yuheng Ren.
\newblock Aria digital twin: A new benchmark dataset for egocentric 3d machine perception, 2023.

\bibitem[Shi et~al.(2023)Shi, Wu, Wu, Liu, Zhao, Feng, Liu, Zhang, Zhang, Zhou, Ding, and Wang]{shi_2023_gir}
Yahao Shi, Yanmin Wu, Chenming Wu, Xing Liu, Chen Zhao, Haocheng Feng, Jingtuo Liu, Liangjun Zhang, Jian Zhang, Bin Zhou, Errui Ding, and Jingdong Wang.
\newblock Gir: 3d gaussian inverse rendering for relightable scene factorization, 2023.

\bibitem[Srinivasan et~al.(2021)Srinivasan, Deng, Zhang, Tancik, Mildenhall, and Barron]{srinivasa_2021_nerv}
Pratul~P. Srinivasan, Boyang Deng, Xiuming Zhang, Matthew Tancik, Ben Mildenhall, and Jonathan~T. Barron.
\newblock Nerv: Neural reflectance and visibility fields for relighting and view synthesis.
\newblock In \emph{CVPR}, pages 7495--7504, New York, NY, USA, 2021. IEEE.

\bibitem[Wang et~al.(2004)Wang, Bovik, Sheikh, and Simoncelli]{wang_2004_ssim}
Zhou Wang, A.C. Bovik, H.R. Sheikh, and E.P. Simoncelli.
\newblock Image quality assessment: from error visibility to structural similarity.
\newblock \emph{IEEE TIP}, 13\penalty0 (4):\penalty0 600--612, 2004.

\bibitem[Yang et~al.(2023)Yang, Chen, Gao, Yuan, Wu, Zhou, and Jin]{yang_2023_sireir}
Ziyi Yang, Yanzhen Chen, Xinyu Gao, Yazhen Yuan, Yu Wu, Xiaowei Zhou, and Xiaogang Jin.
\newblock Sire-ir: Inverse rendering for brdf reconstruction with shadow and illumination removal in high-illuminance scenes, 2023.

\bibitem[Yao et~al.(2022)Yao, Zhang, Liu, Qu, Fang, McKinnon, Tsin, and Quan]{yao2022neilf}
Yao Yao, Jingyang Zhang, Jingbo Liu, Yihang Qu, Tian Fang, David McKinnon, Yanghai Tsin, and Long Quan.
\newblock Neilf: Neural incident light field for physically-based material estimation.
\newblock In \emph{European Conference on Computer Vision (ECCV)}, 2022.

\bibitem[Yariv et~al.(2021)Yariv, Gu, Kasten, and Lipman]{yariv_2021_volsdf}
Lior Yariv, Jiatao Gu, Yoni Kasten, and Yaron Lipman.
\newblock Volume rendering of neural implicit surfaces.
\newblock In \emph{NeurIPS}, pages 4805--4815, Red Hook, NY, USA, 2021. Curran Associates, Inc.

\bibitem[Yu et~al.(2024)Yu, Sattler, and Geiger]{Yu2024GOF}
Zehao Yu, Torsten Sattler, and Andreas Geiger.
\newblock Gaussian opacity fields: Efficient adaptive surface reconstruction in unbounded scenes.
\newblock \emph{ACM Transactions on Graphics}, 2024.

\bibitem[Zhang et~al.(2023)Zhang, Yao, Li, Liu, Fang, McKinnon, Tsin, and Quan]{zhang_2023_neilfpp}
Jingyang Zhang, Yao Yao, Shiwei Li, Jingbo Liu, Tian Fang, David McKinnon, Yanghai Tsin, and Long Quan.
\newblock Neilf++: Inter-reflectable light fields for geometry and material estimation.
\newblock In \emph{ICCV}, pages 3601--3610, New York, NY, USA, 2023. IEEE.

\bibitem[Zhang et~al.(2021{\natexlab{a}})Zhang, Luan, Wang, Bala, and Snavely]{zhang_2021_physg}
Kai Zhang, Fujun Luan, Qianqian Wang, Kavita Bala, and Noah Snavely.
\newblock Physg: Inverse rendering with spherical gaussians for physics-based material editing and relighting.
\newblock In \emph{CVPR}, pages 5453--5462, New York, NY, USA, 2021{\natexlab{a}}. IEEE.

\bibitem[Zhang et~al.(2022{\natexlab{a}})Zhang, Luan, Li, and Snavely]{zhang_iron_2022}
Kai Zhang, Fujun Luan, Zhengqi Li, and Noah Snavely.
\newblock Iron: Inverse rendering by optimizing neural sdfs and materials from photometric images.
\newblock In \emph{CVPR}, pages 5565--5574, New York, NY, USA, 2022{\natexlab{a}}. IEEE.

\bibitem[Zhang et~al.(2018)Zhang, Isola, Efros, Shechtman, and Wang]{zhang_2018_lpips}
Richard Zhang, Phillip Isola, Alexei~A. Efros, Eli Shechtman, and Oliver Wang.
\newblock The unreasonable effectiveness of deep features as a perceptual metric.
\newblock In \emph{CVPR}, New York, NY, USA, 2018. IEEE.

\bibitem[Zhang et~al.(2021{\natexlab{b}})Zhang, Srinivasan, Deng, Debevec, Freeman, and Barron]{zhang_2021_nerfactor}
Xiuming Zhang, Pratul~P. Srinivasan, Boyang Deng, Paul Debevec, William~T. Freeman, and Jonathan~T. Barron.
\newblock Nerfactor: neural factorization of shape and reflectance under an unknown illumination.
\newblock \emph{ACM TOG}, 40\penalty0 (6), 2021{\natexlab{b}}.

\bibitem[Zhang et~al.(2022{\natexlab{b}})Zhang, Sun, He, Fu, Jia, and Zhou]{Zhang22_MII}
Yuanqing Zhang, Jiaming Sun, Xingyi He, Huan Fu, Rongfei Jia, and Xiaowei Zhou.
\newblock Modeling indirect illumination for inverse rendering.
\newblock In \emph{2022 IEEE/CVF Conference on Computer Vision and Pattern Recognition (CVPR)}, pages 18622--18631, 2022{\natexlab{b}}.

\bibitem[Zhang et~al.(2022{\natexlab{c}})Zhang, Sun, He, Fu, Jia, and Zhou]{zhang_2022_mii}
Yuanqing Zhang, Jiaming Sun, Xingyi He, Huan Fu, Rongfei Jia, and Xiaowei Zhou.
\newblock Modeling indirect illumination for inverse rendering.
\newblock In \emph{CVPR}, pages 18643--18652, New York, NY, USA, 2022{\natexlab{c}}. IEEE.

\bibitem[Zhu et~al.(2024)Zhu, Wang, and Yang]{zhu2024gs-ror}
Zuo-Liang Zhu, Beibei Wang, and Jian Yang.
\newblock Gs-ror: 3d gaussian splatting for reflective object relighting via sdf priors.
\newblock \emph{arXiv preprint arXiv:2406.18544}, 2024.

\bibitem[Zwicker et~al.(2002)Zwicker, Pfister, van Baar, and Gross]{Zwicker02EWASplatting}
M. Zwicker, H. Pfister, J. van Baar, and M. Gross.
\newblock Ewa splatting.
\newblock \emph{IEEE Transactions on Visualization and Computer Graphics}, 8\penalty0 (3):\penalty0 223--238, 2002.

\end{thebibliography}
}


\end{document}